\documentclass[a4paper,fleqn]{cas-dc}

\usepackage[authoryear,longnamesfirst]{natbib}
\usepackage{graphicx}
\usepackage{subcaption}
\usepackage{fancyhdr}

\def\tsc#1{\csdef{#1}{\textsc{\lowercase{#1}}\xspace}}
\tsc{WGM}
\tsc{QE}
\tsc{EP}
\tsc{PMS}
\tsc{BEC}
\tsc{DE}

\fancypagestyle{plain}{
  \fancyhf{}
}

\begin{document}
\let\WriteBookmarks\relax
\def\floatpagepagefraction{1}
\def\textpagefraction{.001}




\shorttitle{Enhancing Wrist Fracture Detection with Single-stage detection}

\title [mode = title]{Enhancing Wrist Fracture Detection with YOLO: Analysis of State-of-the-art Single-stage Detection Models}



%
\author[1]{Ammar Ahmed}[type=author,
      orcid=0009-0003-9984-4819,
      ]
\author[2]{Ali Shariq Imran}
\author[1]{Abdul Manaf}
\author[3]{Zenun Kastrati}
\author[1]{Sher Muhammad Daudpota}

\affiliation[1]{  organization={Dept. of Computer Science, Sukkur IBA University}, 
    city={Sukkur},
    postcode={65200}, 
    country={Pakistan}}

\affiliation[2]{organization={Dept. of Computer Science, Norwegian University of Science \& Technology (NTNU)},
    city={ Gj\o vik},
    postcode={2815},
    country={Norway}}

\affiliation[3]{organization={Dept. of Informatics, Linnaeus University},
    city={ Växjö},
    postcode={351 95},
    country={Sweden}}

\begin{abstract}
Diagnosing and treating abnormalities in the wrist, specifically distal radius, and ulna fractures, is a crucial concern among children, adolescents, and young adults, with a higher incidence rate during puberty. However, the scarcity of radiologists and the lack of specialized training among medical professionals pose a significant risk to patient care. This problem is further exacerbated by the rising number of imaging studies and limited access to specialist reporting in certain regions. This highlights the need for innovative solutions to improve the diagnosis and treatment of wrist abnormalities. Automated wrist fracture detection using object detection has shown potential, but current studies mainly use two-stage detection methods with limited evidence for single-stage effectiveness. This study employs state-of-the-art single-stage deep neural network-based detection models YOLOv5, YOLOv6, YOLOv7, and YOLOv8 to detect wrist abnormalities. Through extensive experimentation, we found that these YOLO models outperform the commonly used two-stage detection algorithm, Faster R-CNN, in fracture detection. Additionally, compound-scaled variants of each YOLO model were compared, with YOLOv8m demonstrating a highest fracture detection sensitivity of 0.92 and mean average precision (mAP) of 0.95. On the other hand, YOLOv6m achieved the highest sensitivity across all classes at 0.83. Meanwhile, YOLOv8x recorded the highest mAP of 0.77 for all classes on the GRAZPEDWRI-DX pediatric wrist dataset, highlighting the potential of single-stage models for enhancing pediatric wrist imaging.
\end{abstract}







\maketitle

\section{Introduction}
Wrist abnormalities are a common occurrence in children, adolescents, and young adults. Among them, wrist fractures such as distal radius and ulna fractures are the most common with incidence peaks during puberty \cite{hedstrom_2010, randsborg_et_al_2013, landin_1997, cheng_shen_1993}. Timely evaluation and treatment of these fractures are essential to prevent life-long implications. Digital radiography is a widely used imaging modality to obtain wrist radiographs. While X-ray is often the first and most common imaging modality used for wrist problems, the choice of test depends on the suspected abnormality, clinical presentation, and available resources. If an X-ray doesn't provide a clear diagnosis, other imaging modalities like MRI, CT, or ultrasound may be recommended. The obtained radiographs are then interpreted by surgeons or physicians in training to diagnose wrist abnormalities. However, medical professionals may lack the specialized training to assess these injuries accurately and may rely on radiograph interpretation without the support of an expert radiologist or qualified colleagues \cite{Hallas_Ellingsen_2006}. Studies have shown that diagnostic errors in reading emergency X-rays can reach up to 26\% \cite{Guly_2001, Mounts_2011, Er_2013, Juhl_1990}. This is compounded by the shortage of radiologists even in developed countries \cite{Burki_2018, Rimmer_2017, Makary_Takacs_2022} and limited access to specialist reporting in other parts of the world \cite{rosman_2015} posing a high risk to patient care. The shortage is expected to escalate in the upcoming years due to a growing disparity between the increasing demand for imaging studies and the limited supply of radiology residency positions. The number of imaging studies rises by an average of five percent annually, while the number of radiology residency positions only grows by two percent. \cite{marlow_et_al_2019}. While imaging modalities such as MRI, CT, and ultrasound can further assist in the diagnosis of wrist abnormalities, some fractures may still be occult \cite{fotiadou_patel_morgan_karantanas, neubauer_2016}.

Recent advances in computer vision, more specifically, object detection have shown promising results in medical settings. Some of the positive results of detecting pathologies in trauma X-rays were recently published \cite{adams2020ai, tanzi2020fracture, choi2020dual}. 
In recent years, significant progress has been made in the development of object detection algorithms, leading to their widespread adoption in the medical community. An earlier approach called the sliding window approach \cite{lampert2008beyond} for object detection involved dividing an image into a grid of overlapping regions and then classifying each region as containing the object of interest or not. Key implementations of this method include cascade classifiers that employ LBP (Local Binary Patterns) or Haar-like features. These classifiers are trained using positive examples of a specific object set against random negative images of the same size. Once optimized, the classifier can accurately identify the target object within a specific section of an image. To detect the object throughout the whole image, the classifier systematically examines each segment. It's essential to differentiate between LBP and Haar-like features. LBP characterizes the local texture of an image by comparing a pixel to its neighboring ones, while Haar-like features measure differences in pixel intensities within neighboring rectangular areas. There are several disadvantages of the sliding window approach, one of them being that it is computationally expensive as a large number of regions need to be classified. To address these issues, region-based methods were invented. The main idea behind these methods was to generate candidate object regions and classify only those regions as containing the object of interest or not.

Another method developed as an improvement over the sliding window approach was the single-stage detection method which has gained popularity in recent years due to its efficiency and good performance. This approach uses a single forward propagation through the network to predict bounding boxes and class probabilities, eliminating the need to generate candidate object regions, and making it faster than region-based approaches. While two-stage detection generates candidate regions in the first stage and refines them in the second stage at the cost of speed and computational efficiency, single-stage detection provides a balance between speed and accuracy by predicting final results in a single pass through the network.

Two-stage detection has been the most widely used approach for detecting wrist abnormalities in recent years. However, there has been limited research on the effectiveness of single-stage detectors in detecting various abnormalities in the wrist, including fractures.
In this study, we focus on the effectiveness of SOTA single-stage detectors in detecting wrist abnormalities. Additionally, this study is unique in its use of a large, comprehensively annotated dataset called GRAZPEDWRI-DX presented in a recent publication \cite{Nagy2022}. The characteristics and complexity of the dataset are discussed in section \ref{sec:dataset}. 

Wrist fractures represent just one of several typical wrist abnormalities, other prevalent conditions include Carpal Tunnel Syndrome (CTS), Ganglion Cysts, Osteoarthritis, Tendinitis, as well as Sprains and Strains. Within the dataset that we use, the distinct objects are categorized as fracture, periostealreaction, metal, pronatorsign, softtissue, bone-anomaly, bonelesion, and foreignbody. It's crucial to understand that our primary goal is to detect these specific objects rather than diagnose the overarching abnormalities. In our context, the presence of these objects (including fractures) in the wrist can be considered as 'abnormal'. Moreover, the presence of objects other than fractures may suggest another associated wrist abnormality. For instance, soft tissue presence might be indicative of CTS or a ganglion cyst. In CTS, swelling of the synovial tissue that lines the tendons in the carpal tunnel may be observable. Conversely, a ganglion cyst manifests as a soft tissue structure. The term 'bone lesion' denotes an anomalous area within the bone, severe sprains can involve avulsion fractures where a fragment of bone is pulled away by the ligament.

\subsection{Study Objective \& Research Questions}
The primary objective of this study is to test the effectiveness of the state-of-the-art YOLO detection models, YOLOv5, YOLOv6, YOLOv7, and YOLOv8 on a comprehensively annotated dataset "GRAZPEDWRI-DX" recently released to the public. We compare the performances of all variants within each YOLO model employed to see whether the use of a compound-scaled version of the same architecture improves its performance. Moreover, this study also investigates how effective these single-stage detection methods are in detecting fractures compared to a two-stage detection method widely used in the past. In addition to conducting object detection across multiple classes, we also evaluate the performance of a conventional CNN in binary classification, specifically in distinguishing between fractures and non-fractures. We hypothesize that fractures in the near vicinity of the wrist in pediatric X-ray images can be detected efficiently using YOLO models proposed by \cite{ultralytics_yolov5_2022}, \cite{Li2022}, \cite{wang_bochkovskiy_liao_2022}, and \cite{YOLOv8} respectively. 


We analyze the potential of utilizing object detection techniques in answering the following research questions (RQ):
\begin{enumerate}
   \item To what extent do state-of-the-art YOLO object detection models effectively detect fractures in the vicinity of the wrist in pediatric X-ray images?
   \item In the analysis of wrist images, do the single-stage detection models outperform a two-stage detection model widely used in the past?
   \item Does the use of compound scaled variants within each YOLO algorithm improve its performance in detecting fractures?
   \item To what extent can the YOLO surpass conventional CNN architecture and DenseNets in terms of sensitivity in fracture recognition?
\end{enumerate}
\subsection{Contribution}
The major contributions of this article are as follows:
\begin{itemize}
    \item A thorough performance assessment of SOTA YOLO detection models on the newly released GRAZPEDWRI-DX dataset, a large and diverse set of pediatric X-ray images. To the best of our knowledge, this is the first study of its kind.
    \item An in-depth comparison of the performance of various variants within each YOLO model utilized.
    \item Achieved state-of-the-art mean average precision (mAP) score on the GRAZPEDWRI-DX dataset.
    \item A detailed performance analysis of single-stage detection models in comparison to the widely-used two-stage detection model, Faster R-CNN.
\end{itemize}

\section{Related Work}
Fracture detection is a crucial aspect in the field of wrist trauma, and computer vision techniques have played a significant role in advancing the research in this area. This section provides a comprehensive overview of the existing studies on fracture detection and highlights the key findings. The studies are divided into two subheadings: "Two-stage detection" and "One-stage detection". The first subheading covers studies that have used two-stage detection techniques, while the second subheading focuses on studies that have only employed single-stage detection algorithms.

\subsection{Two-stage detection}
The detection of bone abnormalities, including fracture detection, has been widely studied in the literature, mainly using two-stage detection algorithms. For instance, In a study by \cite{Yahalomi2018}, a Faster R-CNN model utilizing Visual Geometry Group (VGG16) was applied to identify distal radius fractures in anteroposterior wrist X-ray images. The model achieved a mAP of 0.87 when tested on a set of 1,312 images. It should be noted that the initial dataset consisted of only 95 anteroposterior images, with and without fractures, which were then augmented for training as well as for testing.

\cite{thian2019convolutional} developed two separate Faster R-CNN models with Inception-ResNet for frontal and lateral projections of wrist images. The models were trained on 6,515 and 6,537 images of frontal and lateral projections, respectively. The frontal model detected 91\% of fractures, with a specificity of 0.83 and a sensitivity of 0.96. The lateral model detected 96\% of fractures, with a specificity of 0.86 and a sensitivity of 0.97. Both models had a high area under the receiver operating characteristic curve (AUC-ROC) values, with the frontal model having 0.92 and the lateral model having 0.93. The overall per-study specificity was 0.73, sensitivity was 0.98, and AUC was 0.89. 

\cite{Guan2020} used a two-stage R-CNN method to achieve an average precision (AP) of 0.62 on approximately 4,000 X-ray images of arm fractures in musculoskeletal radiographs, MURA dataset. \cite{wang_yao_zhang_guan_wang_2021} developed a two-stage R-CNN network called ParallelNet, with a TripleNet backbone network, for fracture detection in a dataset of 3,842 thigh fracture X-ray images, achieving an AP of 0.88 at an Intersection over Union (IoU) threshold of 0.5.   

\cite{Qi2020GroundTruth} used a Faster R-CNN model with an anchor-based approach, combined with a multi-resolution Feature Pyramid Network (FPN) and a ResNet50 backbone network. They tested the model on 2333 X-ray images of different types of femoral fractures and obtained a mAP score of 0.69.

\cite{raisuddin2021} developed a deep learning-based pipeline called DeepWrist for detecting distal radius fractures. The model was trained on a dataset of 1946 wrist studies and was evaluated on two test sets. The first test set, comprising 207 cases, resulted in an AP score of 0.99, while the second test set, comprising 105 challenging cases, resulted in an AP of 0.64. The model generated heatmaps to indicate the probability of a fracture near the vicinity of the wrist but did not provide a bounding box or polygon to clearly locate the fracture. The study was limited by the use of a small dataset with a disproportionate number of challenging cases. 

\cite{ma2021bone} in their study, first classified the images in the Radiopaedia dataset into the fracture and non-fracture categories using CrackNet. After this, they utilized Faster R-CNN for fracture detection on the 1052 bone images in the dataset. With an accuracy of 0.88, a recall of 0.88, and a precision of 0.89, they demonstrated the usefulness of the proposed approach. \cite{wu2021feature} applied a Feature Ambiguity Mitigate Operator model along with ResNeXt101 and a FPN to identify fractures in a collection of 9040 radiographs of various body parts, including the hand, wrist, pelvic, knee, ankle, foot, and shoulder. They accomplished an AP of 0.77.

\cite{Xue2021} proposed a guided anchoring method (GA) for fracture detection in hand X-ray images using the Faster R-CNN model, which was used to forecast the position of fractures using proposal regions that were refined using the GA module’s learnable and flexible anchors. They evaluated the method on 3067 images and achieved an AP score of 0.71. 

\cite{hardalac2022fracture} conducted 20 fracture detection experiments using a dataset of wrist X-ray images from Gazi University Hospital. To improve the results, they developed an ensemble model by combining five different models, named WFD-C. Out of the 26 models evaluated for fracture detection, the WFD-C model achieved the highest average precision of 0.86. This study utilized both two-stage and single-stage detection methods. The two-stage models employed were Dynamic R-CNN, Faster R-CNN, and SABL and DCN models based on Faster R-CNN. Meanwhile, the single-stage models used were PAA, FSAF, RetinaNet and RegNet, SABL, and Libra.

\cite{Joshi2022} employed transfer learning with a modified Mask R-CNN to detect and segment fractures using two datasets: a surface crack image dataset of 3000 images and a wrist fracture dataset of 315 images. They first trained the model on the surface crack dataset and then fine-tuned it on the wrist fracture dataset. They achieved an average precision of 92.3\% for detection and 0.78 for segmentation on a 0.5 scale, 0.79 for detection, and 0.52 for segmentation on a strict 0.75 scale.

\subsection{One-stage detection}
Very few studies have been conducted demonstrating the performance of one-stage detectors in the area of wrist trauma and fracture detection. In the study by \cite{Sha_Wu_Yu_2020a}, a YOLOv2 model was used to detect fractures in a dataset of 5134 spinal CT images, resulting in a mAP of 0.75. In another research by the same authors \cite{Sha_Wu_Yu_2020b}, a Faster R-CNN model was applied to the same dataset, yielding an mAP of 0.73.

A recent study by \cite{hrzic2022fracture} compared the performance of the YOLOv4 object detection model \cite{bochkovskiy2020} to that of the U-Net segmentation model proposed by \cite{Lindsey2018} and a group of radiologists on the "GRAZPEDWRI-DX" dataset. The authors trained two YOLOv4 models for this study: one for identifying the most probable fractured object in an image and the other for counting the number of fractures present in an image. The first YOLOv4 model achieved high performance, with an AUC-ROC of 0.90 and an F1-score of 0.90, while the second YOLOv4 model achieved an AUC-ROC of 0.90 and an F1-score of 0.96. These results demonstrate the superior performance of YOLOv4 in comparison to traditional methods for fracture detection.

The "GRAZPEDWRI-DX" dataset used in this study was recently published \cite{Nagy2022}. The authors presented the baseline results for the dataset using the COCO pre-trained YOLOv5m variant of YOLOv5. The model was trained on 15,327 (of 20,327) images and tested on 1,000 images. They achieved a mAP of 0.93 for fracture detection and an overall mAP of 0.62 
at an IoU threshold of 0.5.

In conclusion, the literature review shows that the majority of studies on fracture detection have utilized the two-stage detection approach. Additionally, the datasets utilized in these studies tend to be limited in size in comparison to the dataset used in our study. This study builds upon the work of studies \cite{hrzic2022fracture} and \cite{Nagy2022} by conducting a comprehensive comparative study between the state-of-the-art single-stage detection algorithms (YOLOv5, YOLOv6, YOLOv7, and YOLOv8) and a widely used two-stage model Faster R-CNN. The results of this study provide valuable insights into the performance of these algorithms and contribute to the ongoing research in the field of wrist trauma and fracture detection.

\section{Material \& Methods}
\subsection{Research Design}
\begin{figure}[!h]
\centering
\includegraphics[width=1\linewidth, height=7cm]{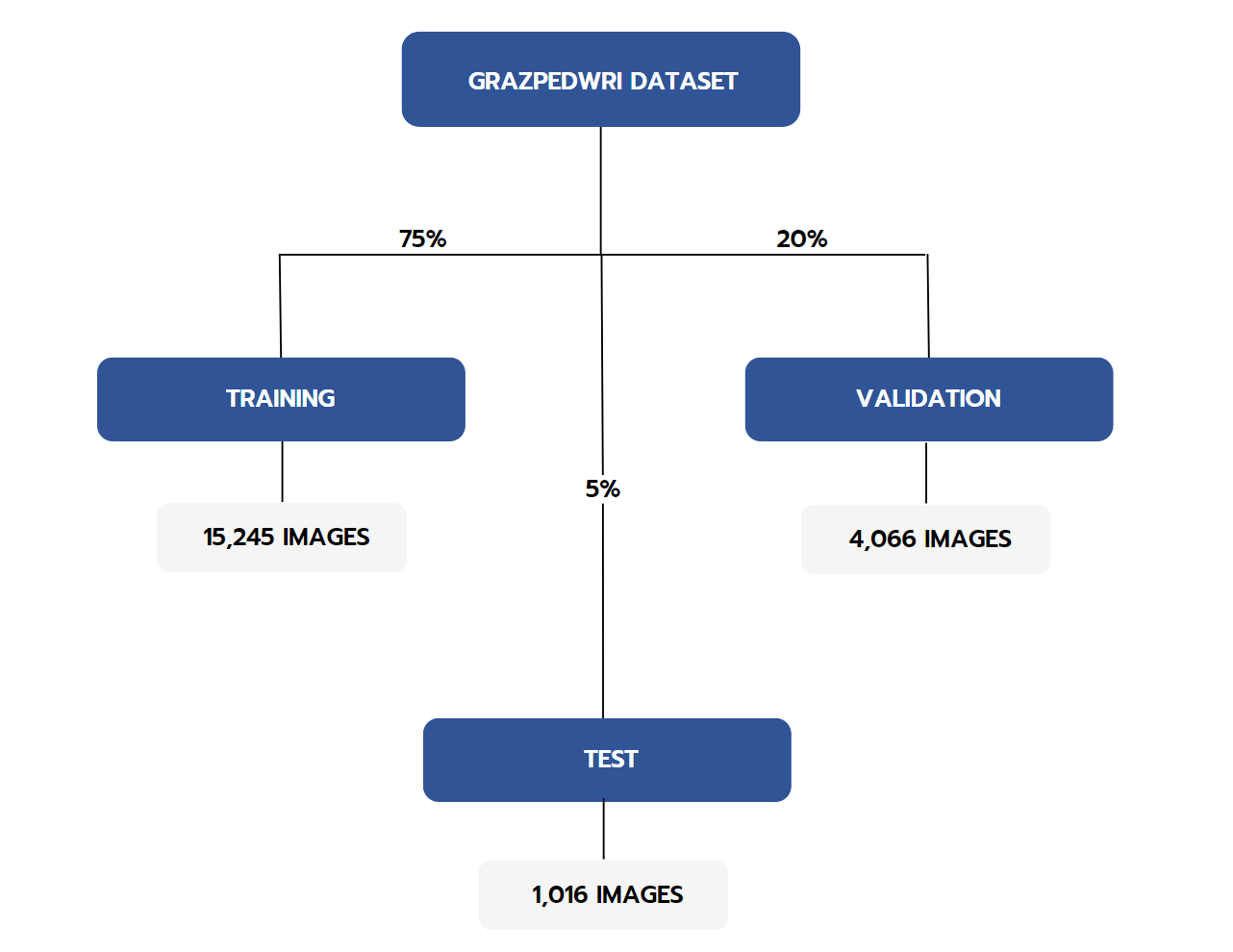}
\caption{Dataset split into training, validation, and test sets.}
\label{fig1:clssplit}
\end{figure}

A quantitative (experimental) study is conducted using data from 10,643 wrist radiography studies of 6,091 unique patients collected by the Division of Paediatric Radiology, Department of Radiology, Medical University of Graz, Austria. As shown in Fig. \ref{fig1:clssplit}, the dataset was randomly partitioned into a training set of 15,245, a validation set of 4,066, and a testing set of 1016. In the following subsection, we describe various measurements used to assess the performance of the models.

\subsection{Tools \& Instruments}
Python scripts were used to partition the dataset into training, validation, and testing sets. The deep learning framework PyTorch was used to train object detection models. To visualize, track, and compare model training, we employed the Weights and Biases (WANDB) platform. To take advantage of our system's graphical processing units (GPUs), we utilized CUDA and cuDNN. All training was performed on a Windows PC equipped with an NVIDIA GeForce RTX 2080 SUPER (with 8,192 MB of video memory), an Intel(R) Xeon(R) W-2223 CPU @ 3.60GHz processor, and 64GB of RAM. The Python version used was 3.9.13. 

\subsection{Deep Learning Models For Object Detection}
In this study, we employed 4 single-stage detection models, namely YOLOv5, YOLOv6, YOLOv7, and YOLOv8, as well as a two-stage detection model Faster R-CNN. To further optimize the performance of the single-stage models, we experimented with multiple variants of each YOLO model, ranging from 5 to 7 variants. This resulted in a total of 23 wrist abnormality detection procedures. 

\begin{figure*}
\includegraphics[width=1\textwidth, height=8.5cm]{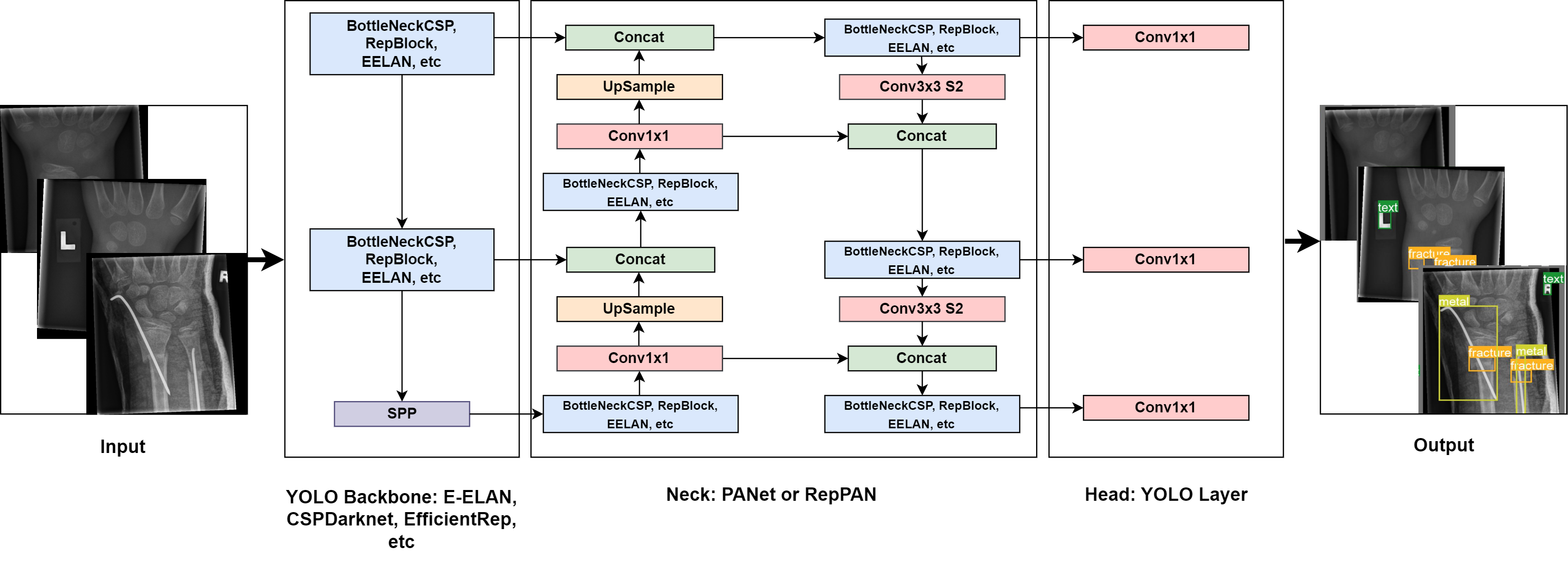}
\caption{YOLO Architecture depicting the input, backbone, neck, head, and the output. }
\label{fig1:customyolo}
\end{figure*}

The YOLO (You Only Look Once) algorithm, initially introduced by \cite{redmon2015} in 2015, is a single-stage object detection approach that uses a single pass of a convolutional neural network to make predictions about the locations of objects in an image, making it faster than other approaches to date. In 2021, YOLOv4 achieved the highest mean average precision on the MS COCO dataset while also being the fastest real-time object detection algorithm \cite{bochkovskiy2020}. Since its initial release, the algorithm has undergone several improvements, with versions ranging from v1 to v8, with each subsequent version offering smaller volume, higher speed, and higher precision. Fig. \ref{fig1:customyolo} illustrates the general structure of YOLO with various backbones used in this study such as CSP, VGG, and EELAN.

\subsubsection{The YOLOv5 Model}

The YOLO framework comprises of three components: the backbone, neck, and head. The backbone extracts image features using the CSPDarknet architecture, known for its superior performance \cite{wang2020cspnet}. We adopted the same architecture in our research. CSPDarknet involves convolution, pooling, and residual connections represented as:
\begin{equation}
F_i = f(F_{i-1}, W_i) + F_{i-1}
\end{equation}
(Where $F_i$ and $F_{i-1}$ are feature maps at $i$-th and $(i-1)$-th layer respectively, $W_i$ represents weights and biases, and $f(\cdot)$ applies convolution and pooling operations). The SPP structure is then used to extract multi-scale features from the CSPDarknet's output:
\begin{equation}
F_{SPP} = g(F_i)
\end{equation}
(Where $F_{SPP}$ denotes multi-scale feature maps, and $g(\cdot)$ performs the SPP operation on $F_i$). The neck component adopts the Path Aggregation Network (PANet) to aggregate backbone features, generating higher-level features for output layers. The head constructs output vectors containing class probabilities, objectness scores, and bounding box coordinates. YOLOv5 encompasses five model variants ("n", "s", "m", "l", and "x"), which are compound-scaled versions of the same architecture. These variants offer varying detection accuracy and performance, achieved by adjusting network depth and layer count.


\subsubsection{The YOLOv6 Model}
YOLOv6 features an anchor-free design and reparameterized Backbone, with VGG and CSP Backbones used in the "n" and "s" variants, and "m", "l" and "l6" variants respectively. This Backbone is referred to as EfficientRep. The Neck, named Rep-PAN, is similar to YOLOv5, but the Head is efficiently decoupled, improving accuracy and reducing computation by not sharing parameters between the classification and detection branches. The YOLOv6 includes five model variants ("n", "s", "m", "l", and "l6"). 


\subsubsection{The YOLOv7 Model}
YOLOv7 comes with several changes, including E-ELAN, which uses expand, shuffle, and merge cardinality to improve network learning without disrupting the gradient path. Other changes include Model Scaling techniques, Re-parameterization planning, and Auxiliary Head Coarse-to-Fine. Model scaling adjusts the width, depth, and resolution of a model to align with specific application requirements. YOLOv7 uses compound scaling to simultaneously scale network depth and width by concatenating layers, maintaining optimal architecture while scaling.

Re-parameterization techniques use gradient flow propagation to identify modules that require averaging weights for robustness. An auxiliary head in the middle of the network improves training but requires a coarse-to-fine approach for efficient supervision. The YOLOv7 model consists of seven variants: "P5" models (v7, v7x, and v7-tiny) and "P6" models (d6, e6, w6, and e6e).

\subsubsection{The YOLOv8 Model}
YOLOv8 is reported to provide significant advancements in object detection when compared to previous YOLO models, particularly in compact versions that are implemented on less powerful hardware. At the time of writing this paper, the architecture of YOLOv8 is not fully disclosed and some of its features are still under development. As of now, it's been confirmed that the system has a new backbone, uses an anchor-free design, has a revamped detection head, and has a newly implemented loss function. We have included the performance of this model on the GRAZPEDWRI-DX dataset as a benchmark for future studies, as further improvements to YOLOv8 may surpass the results obtained in this study. YOLOv8 comes in five versions at the time of release (January 10, 2023), namely, "n", "s", "m", "l", and "x". 



\subsubsection{Faster R-CNN}
\begin{figure*}
\includegraphics[width=1\textwidth, height=7.5cm]{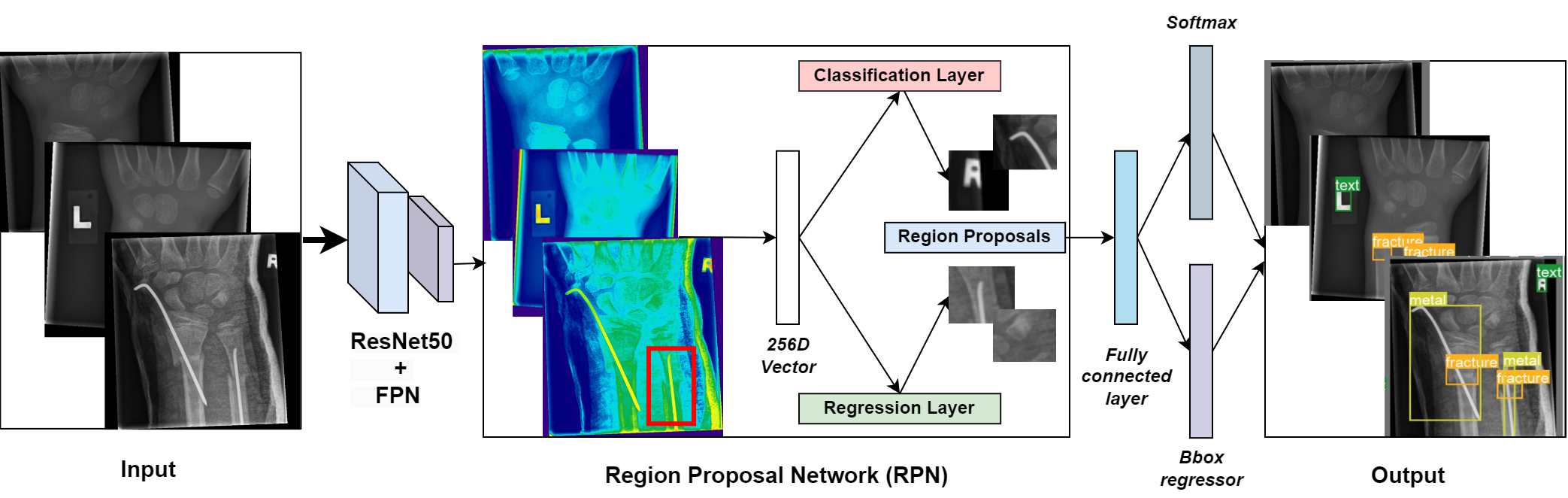}
\caption{Faster R-CNN Pipeline.}
\label{fig1:faster_rcnn_pipeline}
\end{figure*}

The Faster R-CNN model includes a backbone, an RPN (regional proposal network), and a detection network. ResNet50 with FPN is used as the backbone for feature extraction. Anchors with variable sizes and aspect ratios are generated for each feature. The RPN selects appropriate anchor boxes using a classifier that predicts if an anchor box contains an object based on an IoU threshold of 0.5. The regressor predicts offsets for anchor boxes containing objects to fit them tightly to the ground truth labels. Finally, the RoI pooling layer converts variable-sized proposals to a fixed size to run a classifier and regress a bounding box. Fig. \ref{fig1:faster_rcnn_pipeline} illustrates the architecture of Faster R-CNN.


\subsection{Training Details}
In the experimentation of YOLO variants, standard hyperparameters were utilized. The input resolution was fixed at 640 pixels. The optimization algorithm employed was SGD with an initial learning rate $\alpha = 1 \times 10^{-2}$, final learning rate $\alpha_{f} = 1 \times 10^{-2}$ (except for YOLOv7 variants with a final learning rate $\alpha_{f} = 1 \times 10^{-1}$), momentum = 0.937, weight decay = $5 \times 10^{-4}$. Each variant/model underwent 100 epochs of training from scratch and was observed to converge between 90-100 epochs. Every variant was trained with a batch size of 16 except for the "P6" variants of YOLOv7 namely (d6, e6, w6, e6e) which were trained with a batch size of 8 due to computational constraints. 

With Faster R-CNN, the only difference was the learning rate of $\alpha = 1 \times 10^{-3}$, momentum of 0.9 and weight decay of $5 \times 10^{-4}$. All other parameters were the same as YOLO variants. As with YOLO models, the selection of these parameters is not deliberate, they are the default settings. 

All binary classifiers were trained for a maximum of 100 epochs using a batch size of 64. The learning rate was set at $1 \times 10^{-3}$. The Adam optimization algorithm guided the training process. Input images were standardized to a resolution of 224 pixels.

\subsection{Evaluation Metrics: mAP}
\begin{figure}
\includegraphics[width=0.49\textwidth, height=7cm]{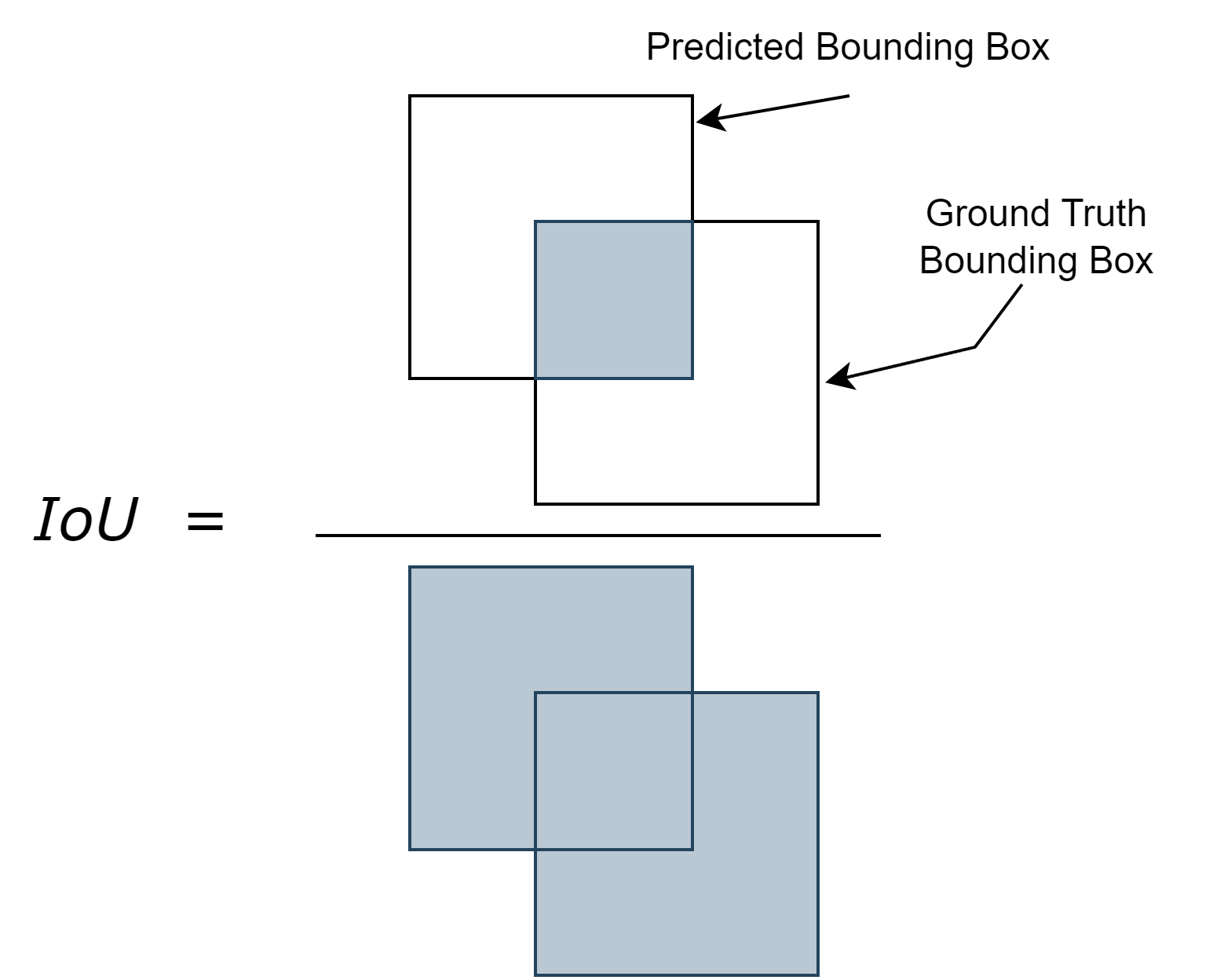}
\caption{Visual illustration of Intersection over Union (\textit{IoU}).}
\label{fig1:iou}
\end{figure}

For the evaluation of object detection, a common way to determine if the predicted location of an object was correct is to find in {\it{Intersection over Union (IoU)}}. It is defined as the ratio of the intersection of the predicted and the ground truth bounding box over the union of the predicted and ground truth bounding box. A visual illustration of \textit{IoU} is presented in Fig. \ref{fig1:iou}. Given the set of predicted bounding boxes \textit{A} for a given image, and the set of ground truth bounding boxes \textit{B} for the same image. The IoU can be computed as:
\begin{equation}
\textit{IoU(A, B)} = \frac{A \cap B}{A \cup B}\text{;} \qquad\text{where } A, B \in [0, 1]
\end{equation}
Commonly, if the $\textit{IoU} > 0.5$, we classify the detection as true positive, otherwise, it is classified as false positive. Given \textit{IoU}, we can compute the number of true positives \textit{TP} and false positives \textit{FP} and compute the Average precision \textit{AP} for each object class \textit{c} as follows:
\begin{equation}
\textit{AP(c)} = \frac{\textit{TP(c)}}{\textit{TP(c) + FP(c)}}
\end{equation}
Finally, after computing \textit{AP} for each object class, we compute the Mean Average Precision \textit{mAP} which is an average of \textit{AP} across all classes \textit{C} under consideration. \textit{mAP} is given as:
\begin{equation}
\textit{mAP} = \frac{1}{C} \sum_{c=1}^{C} \textit{AP(c)}
\end{equation}
\textit{mAP} is the metric that quantifies the performance of object detection algorithms. Thus, the metric $\textit{mAP}_{0.5}$ indicates \textit{mAP} for $\textit{IoU} > 0.5$. This is the \textit{IoU} threshold we will be using to make our assessments of the detection models.

\subsubsection{Sensitivity}
Sensitivity, in the context of our model, pertains to its capacity to accurately recognize true detections among all positive detections within the dataset. Specifically, it gauges the model's ability to correctly identify the presence of a fracture or abnormality. We prioritize this metric due to the potential consequences of false negatives in wrist trauma cases. Failure to detect fractures is a frequent reason for differences in diagnosis between the initial interpretation of X-ray images and the final analysis conducted by certified radiologists. The calculation for sensitivity is as follows:
\begin{equation}
\text{Sensitivity} = \frac{\text{True Positives}}{\text{True Positives} + \text{False Negatives}}
\end{equation}

\section{Dataset}\label{sec:dataset}
The dataset used in this study is called GRAZPEDWRI-DX \textit{for machine learning} presented by the authors in \cite{Nagy2022} and is publicly made available to encourage computer vision research. The dataset contains pediatric wrist radiograph images in PNG format of 6,091 patients (mean age 10.9 years, range 0.2 to 19 years; 2,688 females, 3,402 males, 1 unknown), treated at the Division of Paediatric Radiology, Department of Radiology, Medical University of Graz, Austria. The dataset includes a total of 20,327 wrist images covering lateral and posteroanterior projections. The radiographs were acquired over the span of 10 years between 2008 and 2018 and have been comprehensively annotated between 2018 and 2020 by expert radiologists and various medical students. The annotations were validated by three experienced radiologists as the X-ray images were annotated. This process was repeated until a consensus was met between the annotations and interpretations from three radiologists. We choose to use this dataset in our study for the following reasons: 
\begin{enumerate}
    \item The dataset is quite large consisting of 20,327 labeled and tagged images, making it suitable for various computer vision algorithms

    \item To our knowledge, there are no related pediatric datasets publicly available, with others featuring only binary labels or not as comprehensively labeled as the one we use.
  
    \item To the best of our knowledge, this is the first comprehensive study of the recently released GRAZPEDWRI-DX dataset using state-of-the-art computer vision models YOLOv5, v6, v7 and v8.

    \item It contains diverse images of the early stages of bone growth and organ formation in children. Studying the wrist at this stage offers unique insights into the diagnosis, treatment, and prevention of anomalies that are not possible when studying adult wrists.
\end{enumerate}

\subsection{Analysis of Objects in the Dataset}
\begin{figure}[!h]
\centering
\includegraphics[width=0.8\linewidth]{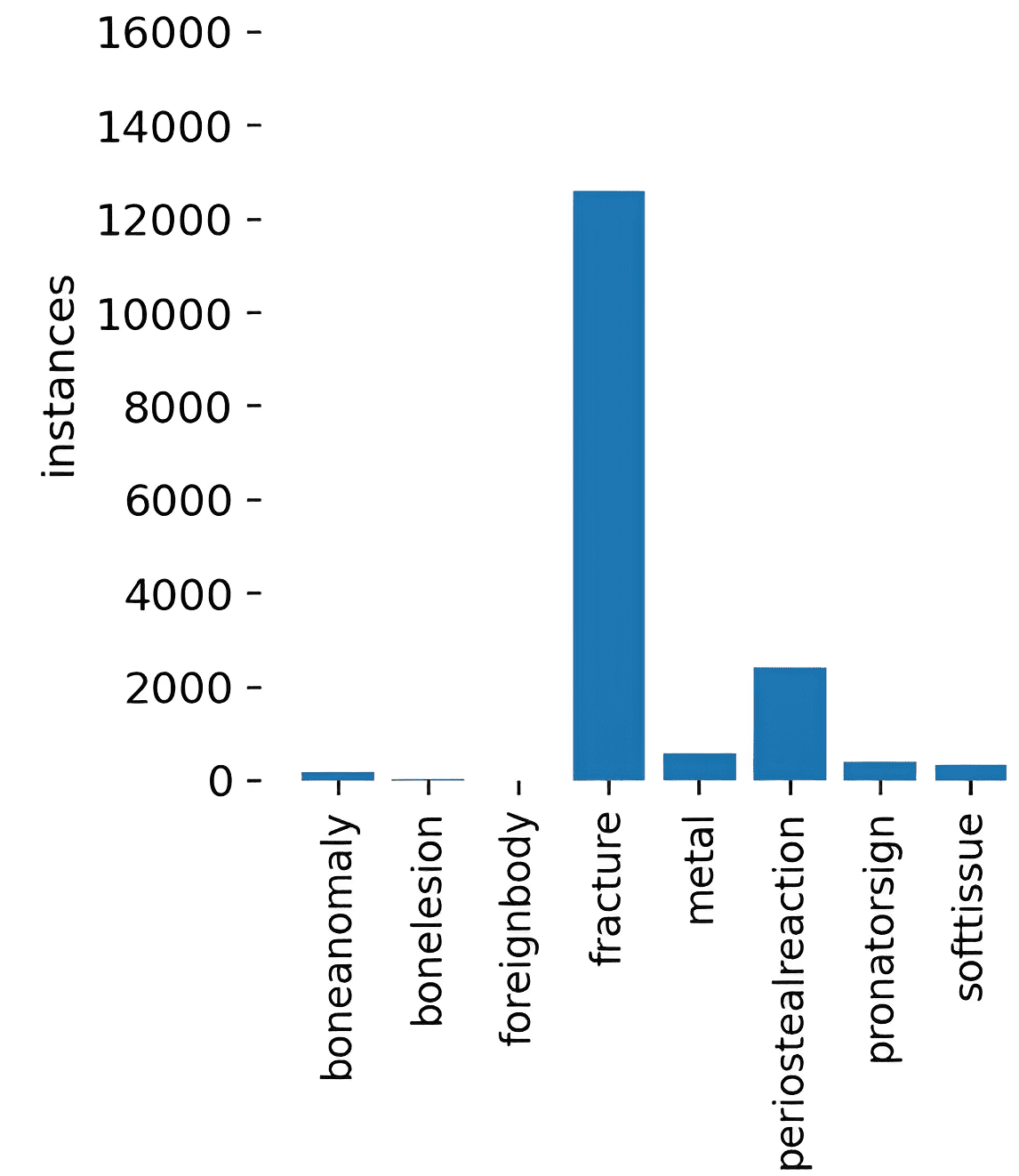}
\caption{Histogram of Class Distribution.}
\label{fig1:hist}
\end{figure}

The dataset includes a total of 9 objects: periosteal reaction, fracture, metal, pronator sign, soft tissue, bone anomaly, bone lesion, foreign body, and text. The object "text" is present in all X-ray images and is used to identify the side of the body (right or left hand) on which the X-ray was taken. The number of objects in the dataset is shown in Table \ref{tab:table1}. The table clearly indicates that the object "fracture" has the most common occurrence in wrist X-rays of GRAZPEDWRI-Dataset. The class "periosteal reaction" has the second largest occurrence followed by the third largest class "metal". Meanwhile, the classes "bone anomaly", "bone lesion", and "foreign body" have the lowest occurrence. Note that this table shows how many X-ray images contain a particular object and not the number of times an object is labeled in the dataset. Additionally, a histogram is shown in Fig. \ref{fig1:hist} visually shows the class distribution.

\begin{table}[width=.9\linewidth,cols=4,pos=h]
\caption{Class Distribution}
    \begin{tabular}{l c c }
    \hline
    Abnormality & Instances & Ratio\\
    \hline 
  Boneanomaly                                                              & 192                                                               & 0.94\%                                                                            \\
  Bonelesion                                                                & 42                                                                & 0.21\%                                                                            \\
  Foreignbody                                                               & 8                                                                 & 0.04\%                                                                            \\
  Fracture                                                                 & 13550                                                             & 66.6\%                                                                           \\
  Metal                                                                     & 708                                                               & 3.48\%                                                                            \\
  Periostealreaction                                                        & 2235                                                              & 11.0\%                                                                           \\
  Pronatorsign                                                              & 566                                                               & 2.78\%                                                                            \\
  Softtissue                                                                & 439                                                               & 2.16\%                     \\\hline                                                      
  \end{tabular}
  \label{tab:table1}
\end{table}

In Table \ref{tab:table2}, we show the number of images in which a particular anomaly occurs only once, twice, or multiple times. The column "Total" represents the total number of images in which a particular anomaly is present.

\begin{table}[width=.9\linewidth,cols=4,pos=h]
\caption{Object Occurrences}\label{tab:table2}
    \begin{tabular}{l c c c c c}
    \hline
    Abnormality & Zero & One & Two & More & Total\\
    \hline 
Fracture     & 6777  & 9212   & 4137  & 201   & 13550 \\
Boneanomaly     & 20135    & 42   & 24   & 126   & 192   \\
Bonelesion    & 20285     & 11    & 8    & 23   & 42  \\
Foreignbody                                           & 20319                                                    & 0                                                       & 0                                                       & 8                                                        & 8                                                             \\
Metal                                                  & 19620                                                                            & 347                                                                             & 219                                                                             & 141                                                                              & 707                                                           \\
Periostealreaction                                     & 18092                                                    & 1273                                                    & 885                                                     & 77                                                       & 2235                                                          \\
Pronatorsign                                           & 19761                                                                            & 456                                                                             & 71                                                                              & 39                                                                               & 566                                                           \\
Softtissue                                             & 19888                                                    & 221                                                     & 82                                                      & 136                                                      & 439  \\\hline                                                        
\end{tabular}

\end{table}

\section{Results \& Discussion}

\begin{figure*}[h!]
    \centering
    \begin{subfigure}{.49\textwidth}
         \centering
         \includegraphics[width=1\linewidth]{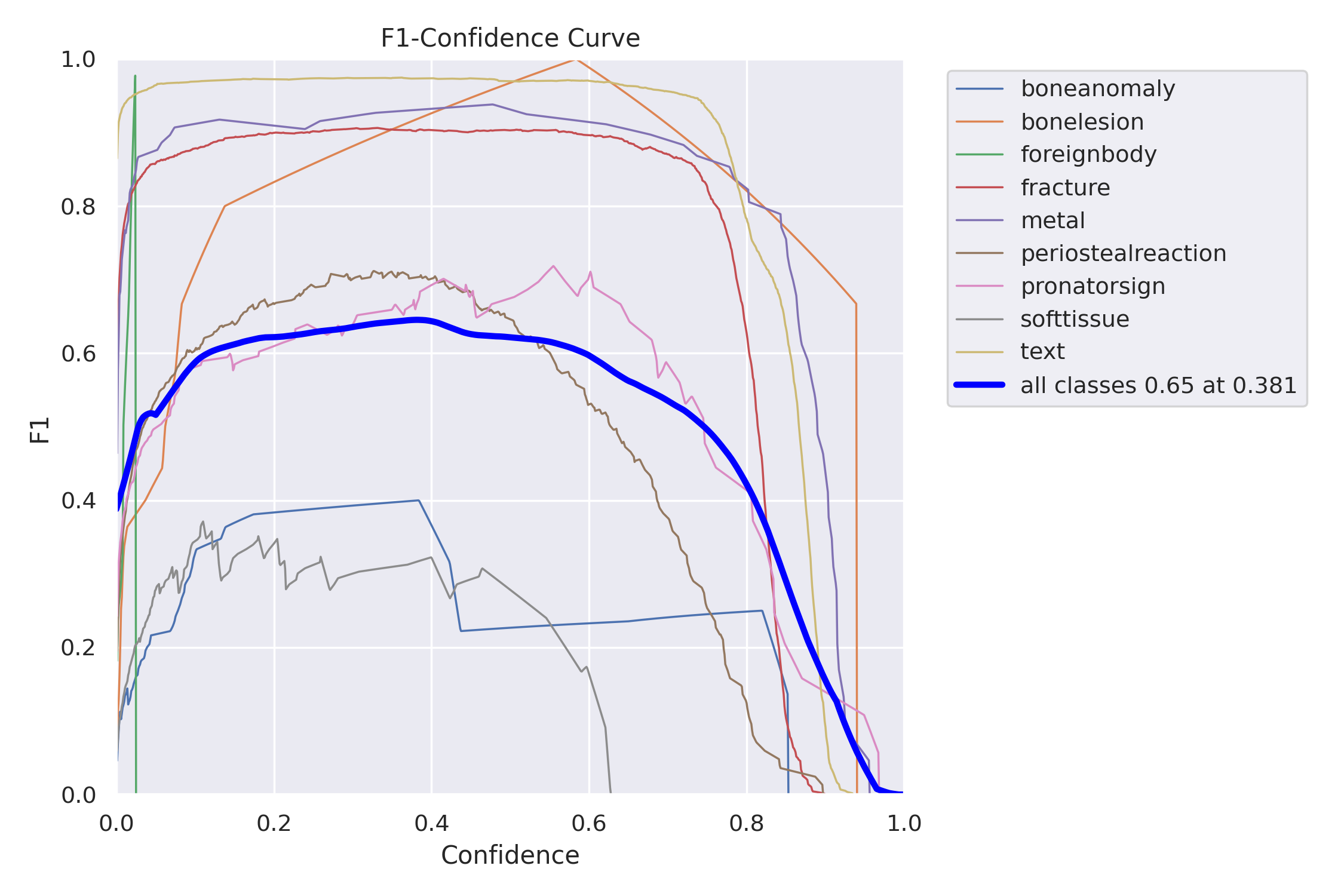}
         \caption{F1 vs. Confidence}
         \label{f1fig1}
    \end{subfigure}
    \hfill
    \begin{subfigure}{0.49\textwidth}
        \centering
        \includegraphics[width=1\linewidth]{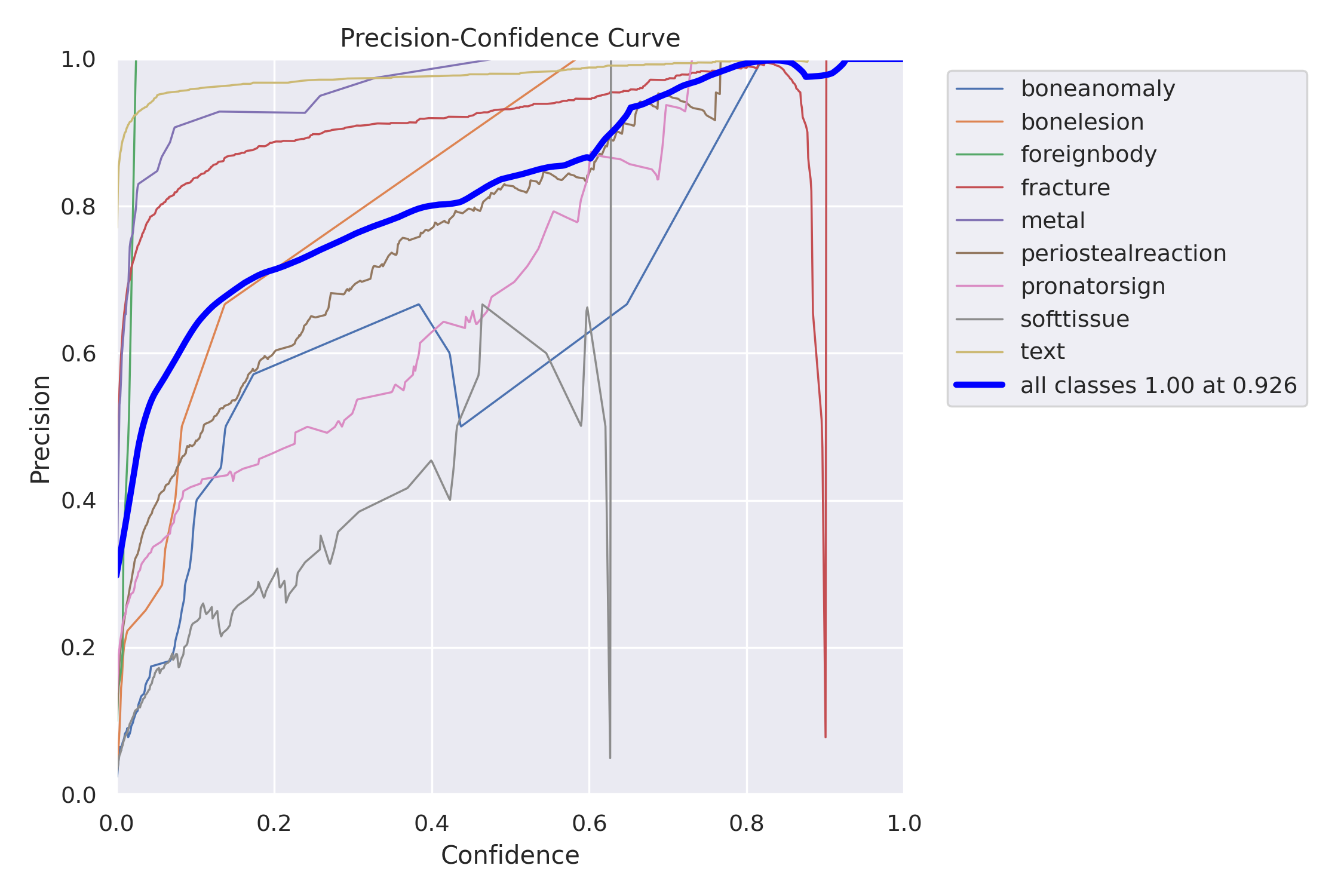}
        \caption{Precision vs. Confidence}
        \label{precfig1}
    \end{subfigure}    
    \hfill
    \vspace{0.5cm}
    \begin{subfigure}{0.49\textwidth}
         \centering
         \includegraphics[width=1\linewidth]{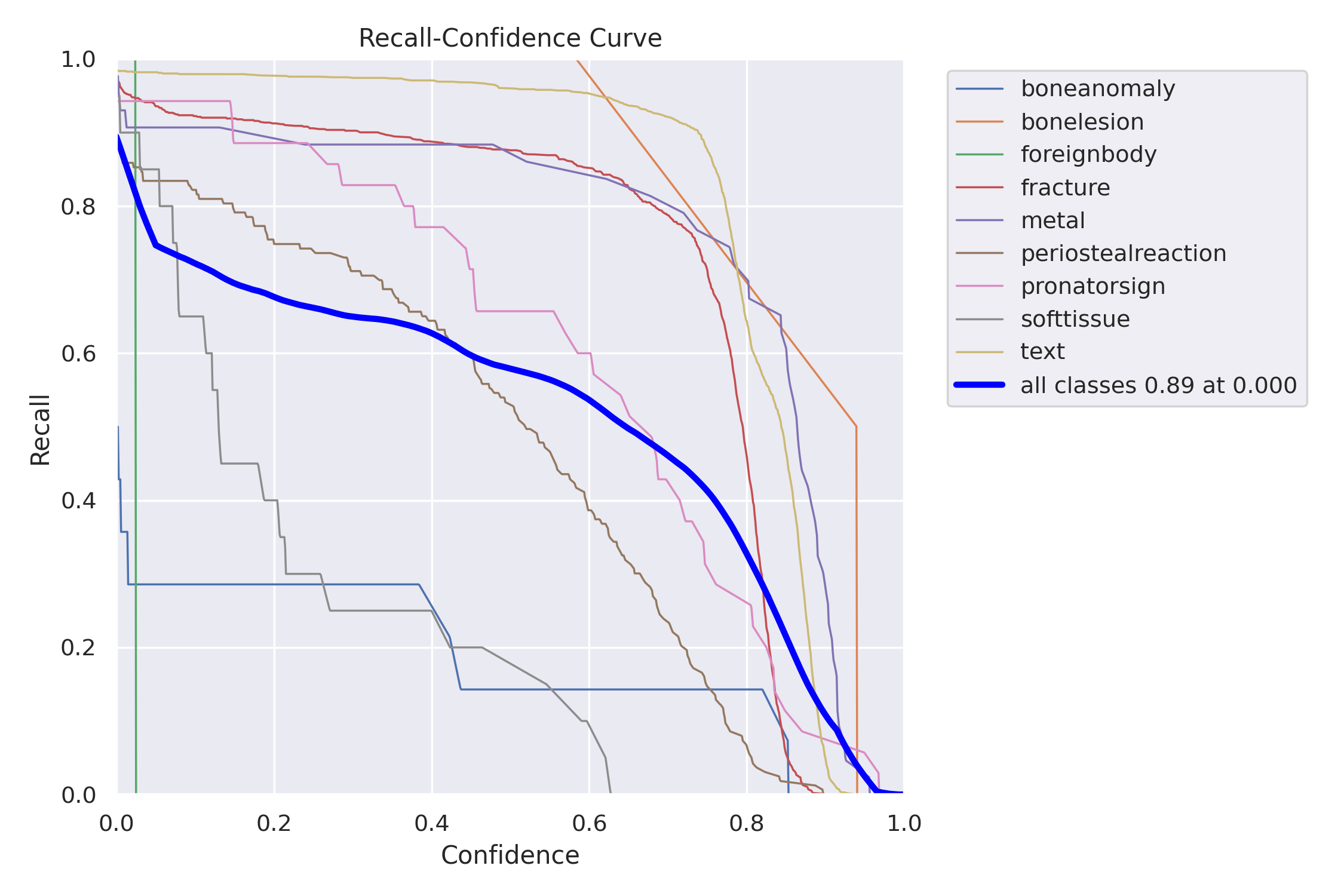}
         \caption{Recall vs. Confidence}
         \label{recfig1}
     \end{subfigure}
     \hfill
    \begin{subfigure}{0.49\textwidth}
        \centering
        \includegraphics[width=1\linewidth]{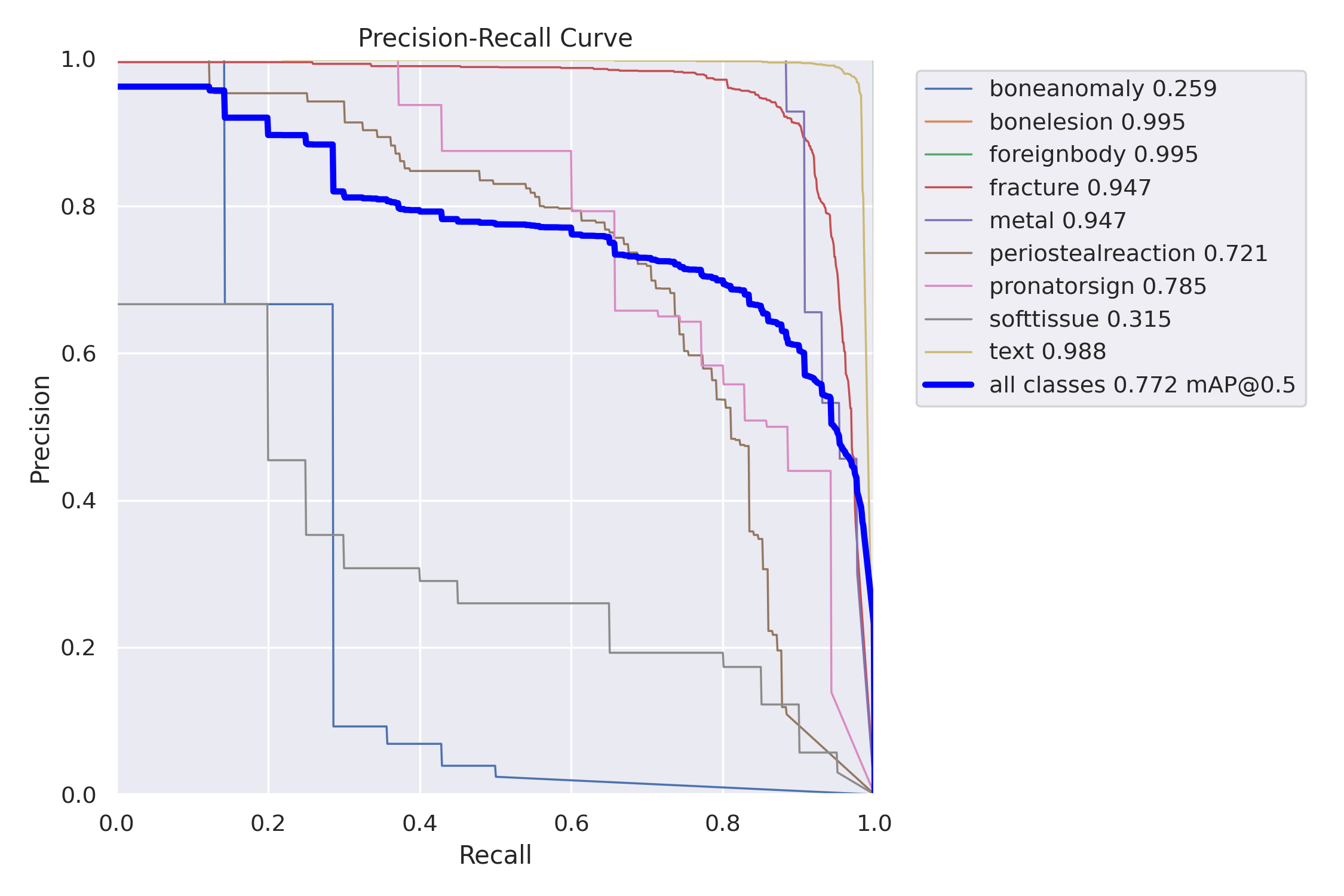}
        \caption{Precision vs. Recall}
        \label{prfig1}
     \end{subfigure}
     \caption{Performance analysis curves (YOLOv8x)}
     \label{fourfig}
\end{figure*}

This section presents a comprehensive analysis of the performance of various models for wrist abnormality detection on the GRAZPEDWRI-DX dataset. A total of 23 detection procedures were conducted using different variants of each YOLO model and a two-stage detection model (Faster R-CNN) on a test set consisting of 1016 randomly selected samples. The performance of each model was evaluated using metrics such as precision, recall, and mean average precision (mAP). We begin by providing a detailed analysis of the variants within each YOLO model. Next, we select the best-performing variant from each YOLO model based on the highest mAP score obtained for the fracture class, as well as across all classes. Finally, we compare these variants to determine the overall best-performing model and evaluate its performance against Faster R-CNN.

The results of YOLOv5 variants are presented in Table \ref{tab:table3} and \ref{tab:table4}, showing the performance of the variants across all classes and on the fracture class, respectively. All values are rounded to two decimal places. The results show that the fractures were detected with the highest mAP of 0.95 at IoU = 0.5, with a precision of 0.92, and a recall of 0.90 by the YOLOv5 variant, YOLOv5l. Additionally, the performance of YOLOv5l appears to be satisfactory across all classes with the mAP score of 0.68 at IoU = 0.5. The variant YOLOv5x seems to perform just as well in terms of mAP obtained for the fracture class. In terms of overall performance across all classes, the highest mAP score achieved was 0.69 by the two YOLOv5 variants "m" and "x". The highest precision obtained across all classes is 0.80 by the variant "m", while the highest recall achieved was 0.66 by the variant "s". It can also be observed from the results shown in Table \ref{tab:table4} that as the complexity of the architecture in YOLOv5 increases, its performance improves.

Table \ref{tab:table5} displays the mAP scores of all YOLOv5 variants at an IoU threshold of 0.5 for all classes present in the GRAZPEDWRI-DX dataset. It is worth noting that these mAP scores are particularly significant as they are calculated at an IoU threshold of 0.5, which is a commonly used threshold in object detection evaluations. These scores are crucial indicators of the performance of the YOLOv5 variants on the GRAZPEDWRI-DX dataset and provide valuable insights into their abilities to detect objects within the various classes present in the GRAZPEDWRI-DX dataset. Upon examination of the Table \ref{tab:table5}, it can be seen that almost all variants of YOLOv5 demonstrate the capability to detect classes that are in the minority, such as bone anomaly, bone lesion, and foreign body, with considerably good mAP scores as seen in Table \ref{tab:table5}. For instance, despite the limited number of instances of the class "Bonelesion" (only 42, as shown in Table \ref{tab:table1}), the four variants of YOLOv5 ("s", "m," "l," and "x") are able to correctly detect it in all instances where it occurs, with the mAP score of 1.00. 

\begin{table}[htbp]
\caption{YOLOv5 Results (Across All Classes)}
\small 
\setlength\tabcolsep{4pt} 
\begin{tabular}{@{}lccccc@{}}
\hline
Model variant & Precision & Recall & mAP@0.5 & mAP@0.5:0.95\\
\hline 
YOLOv5n & 0.77 & 0.52 & 0.59 & 0.34 \\
YOLOv5s & 0.75 & 0.66 &  0.65 & 0.38 \\
YOLOv5m & 0.80 & 0.62 & 0.69 & 0.44 \\
YOLOv5l & 0.76 & 0.61 & 0.68 & 0.43 \\
YOLOv5x & 0.73 & 0.64 & 0.69 & 0.45 \\
\hline
\end{tabular}
\label{tab:table3}
\end{table}

\begin{table}[htbp]
\caption{YOLOv5 Results (Fracture Class)}
\small 
\setlength\tabcolsep{4pt} 
\begin{tabular}{@{}lccccc@{}}
\hline
    Model variant & Precision & Recall & mAP@0.5 & mAP@0.5:0.95\\
    \hline
YOLOv5n & 0.87 & 0.91 & 0.94 & 0.54 \\
YOLOv5s & 0.89 & 0.91 & 0.95 & 0.56 \\
YOLOv5m & 0.91 & 0.90 & 0.94 & 0.56 \\
YOLOv5l & 0.92 & 0.90 & 0.95 & 0.57 \\
YOLOv5x & 0.91 & 0.90 & 0.95 & 0.57 \\\hline
\end{tabular}
\label{tab:table4}
\end{table}


\begin{table*}[h]
\caption{YOLOv5 mAP@0.5 Scores (For All Classes)}
\small 
\setlength\tabcolsep{3.5pt}
    \begin{tabular}{l c c c c c c c c c}
    \hline
    Model variant & Boneanomality & Bonelesion & Foreignbody & Fracture & Metal & Periostealreaction & Pronatorsign & Softtissue & Text\\\hline
YOLOv5n & 0.31 & 0.57 & 0.00 & 0.94 & 0.88 & 0.66 & 0.74 & 0.21 & 0.99\\
YOLOv5s & 0.31 & 1.00 & 0.00 & 0.95 & 0.91 & 0.75 & 0.74 & 0.25 & 0.99\\
YOLOv5m & 0.33 & 1.00 & 0.33 & 0.94 & 0.92 & 0.69 & 0.75 & 0.25 & 0.99\\
YOLOv5l & 0.34 & 1.00 & 0.25 & 0.95 & 0.90 & 0.71 & 0.75 & 0.19 & 0.99\\
YOLOv5x & 0.37 & 1.00 & 0.33 & 0.95 & 0.92 & 0.71 & 0.77 & 0.19 & 0.99\\\hline
\label{tab:table5}
\end{tabular}
\end{table*}

Table \ref{tab:table14} and \ref{tab:table15} present the results of YOLOv6 variants, showcasing their performance on all classes and the fracture class, respectively. Variants "n", "s", and "m" achieved the highest mAP of 0.94 at an IoU threshold of 0.5 for detecting fractures. Variants "n", "m", and "l" displayed the highest precision for the fracture class with a value of 0.94, while variant "s" had the highest recall of 0.89. In terms of overall performance across all classes, the highest mAP score of 0.64 at an IoU threshold of 0.5 was obtained by variants "m" and "l", with variant "l" achieving the highest precision of 0.60 and variant "m" having the highest recall of 0.83. 

Table \ref{tab:table16} illustrates that YOLOv6 variants, similar to YOLOv5 variants, exhibit the ability to detect minority classes. However, Table \ref{tab:table15} reveals that, unlike YOLOv5, as the complexity of the model increases from variant "m" to "l" and then to "l6", the mAP score decreases, indicating that complexity beyond variant "m" results in decreased performance. This trend is also observed in Table \ref{tab:table14}, where increasing complexity from variant "l" to "l6" results in decreased performance across all classes.

\begin{table}[htbp]
\caption{YOLOv6 Results (Across All Classes)}
\small 
\setlength\tabcolsep{4pt} 
\begin{tabular}{@{}lccccc@{}}
\hline
    Model variant & Precision & Recall & mAP@0.5 & mAP@0.5:0.95\\
    \hline
YOLOv6n & 0.50 & 0.73 & 0.51 & 0.31 \\
YOLOv6s & 0.51 & 0.82 &  0.62 & 0.37 \\
YOLOv6m & 0.59 & 0.83 & 0.64 & 0.36 \\
YOLOv6l & 0.60 & 0.80 & 0.64 & 0.41 \\
YOLOv6l6 & 0.49 & 0.77 & 0.52 & 0.31 \\\hline
\end{tabular}
\label{tab:table14}
\end{table}

\begin{table}[htbp]
\caption{YOLOv6 Results (Fracture Class)}
\small 
\setlength\tabcolsep{4pt} 
\begin{tabular}{@{}lccccc@{}}
\hline
    Model variant & Precision & Recall & mAP@0.5 & mAP@0.5:0.95\\
    \hline
YOLOv6n & 0.94 & 0.86 & 0.94 & 0.55 \\
YOLOv6s & 0.92 & 0.89 & 0.94 & 0.54 \\
YOLOv6m & 0.94 & 0.87 & 0.94 & 0.55 \\
YOLOv6l & 0.94 & 0.87 & 0.93 & 0.53 \\
YOLOv6l6 & 0.91 & 0.86 & 0.92 & 0.53 \\\hline
\end{tabular}
\label{tab:table15}
\end{table}

\begin{table*}[!htb]
\caption{YOLOv6 mAP@0.5 Scores (For All Classes)}
\small 
\setlength\tabcolsep{3.5pt}
    \begin{tabular}{l c c c c c c c c c}
    \hline
    Model variant & Boneanomality & Bonelesion & Foreignbody & Fracture & Metal & Periostealreaction & Pronatorsign & Softtissue & Text\\\hline
YOLOv6n & 0.10 & 0.01 & 0.00 & 0.94 & 0.84 & 0.73 & 0.76 & 0.29 & 0.98\\
YOLOv6s & 0.15 & 0.54 & 0.33 & 0.94 & 0.91 & 0.72 & 0.76 & 0.21 & 0.98\\
YOLOv6m & 0.10 & 0.10 & 1.00 & 0.94 & 0.87 & 0.75 & 0.76 & 0.31 & 0.98\\
YOLOv6l & 0.10 & 0.10 & 1.00 & 0.93 & 0.93 & 0.71 & 0.77 & 0.25 & 0.98\\
YOLOv6l6 & 0.13 & 0.10 & 0.00 & 0.92 & 0.90 & 0.67 & 0.76 & 0.22 & 0.98\\\hline
\label{tab:table16}
\end{tabular}
\end{table*}

The performance of YOLOv7 variants on both across classes and the fracture class is presented in Tables \ref{tab:table6} and \ref{tab:table7}, respectively. The results indicate that the second variant of the YOLOv7 model exhibits the highest mean average precision (mAP) of 0.94 at an intersection over union (IoU) threshold of 0.5, with a precision of 0.86 and recall of 0.91 for detecting fractures. This variant also demonstrates superior performance across all classes with a mAP of 0.61 at an IoU of 0.5, a precision of 0.79, and a recall of 0.54. The variant YOLOv7x seems to perform just as well in terms of mAP obtained for the fracture class but has a lower mAP score compared to the second variant across all classes. Additionally, it can be observed from our experiments that, in contrast to YOLO5, increasing the complexity of the YOLOv7 architecture, in terms of depth and number of layers, hurts its performance in detecting wrist abnormalities. The only exception to this trend is the increase in performance observed when comparing the smaller variant "YOLOv7-Tiny" to the slightly larger variant "YOLOv7". The "YOLOv7-Tiny" achieved mAP of 0.5 at IoU=0.5, but the "YOLOv7" variant showed an improvement of 0.11 across all classes. Additionally, when focusing on the specific class of fractures, an improvement of 0.01 in the mAP score was observed, suggesting that there is an optimal balance of complexity and performance for this model.  The decline in performance for YOLOv7's "P6" models, specifically "W6", "E6", "D6", and "E6E", compared to the "P5" models may be attributed to the reduced image resolution. However, the results across all classes indicate that even with this resolution, the performance of "P6" models either decreases or does not improve at all.

It is worth noting that rare classes such as bone anomaly, bone lesion, and foreign body have a very low mAP score and are sometimes not detected at all, as shown in Table \ref{tab:table9}. However, the second variant of YOLOv7 is the only variant able to detect all the minority classes such as "bone anomaly", "bone lesion", and "foreign body". 

\begin{table}[htbp]
\caption{YOLOv7 Results (Across All Classes)}
\small 
\setlength\tabcolsep{4pt} 
\begin{tabular}{@{}lccccc@{}}
\hline
    Model variant & Precision & Recall & mAP@0.5 & mAP@0.5:0.95\\
    \hline
YOLOv7-Tiny & 0.59 & 0.52 & 0.50 & 0.28\\
YOLOv7 & 0.79 & 0.54 & 0.61 & 0.39 \\
YOLOv7x & 0.68 & 0.49 & 0.53 & 0.32 \\
YOLOv7-W6 & 0.53 & 0.43 & 0.44 & 0.24 \\
YOLOv7-E6 & 0.56 & 0.38 & 0.40 & 0.21 \\
YOLOv7-D6 & 0.50 & 0.45 & 0.42 & 0.24 \\
YOLOv7-E6E & 0.75 & 0.45 & 0.44 & 0.25 \\\hline
\end{tabular}
\label{tab:table6}
\end{table}

\begin{table}[htbp]
\caption{YOLOv7 Results (Fracture Class)}
 \small 
\setlength\tabcolsep{4pt} 
\begin{tabular}{@{}lccccc@{}}
\hline
    Model variant & Precision & Recall & mAP@0.5 & mAP@0.5:0.95\\
    \hline
YOLOv7-Tiny & 0.79 & 0.91 & 0.93 & 0.53\\
YOLOv7 & 0.86 & 0.91 & 0.94 & 0.55 \\
YOLOv7x & 0.85 & 0.90 & 0.94 & 0.54 \\
YOLOv7-W6 & 0.72 & 0.89 & 0.90 & 0.50 \\
YOLOv7-E6 & 0.55 & 0.84 & 0.83 & 0.44 \\
YOLOv7-D6 & 0.62 & 0.89 & 0.89 & 0.49 \\
YOLOv7-E6E & 0.74 & 0.89 & 0.91 & 0.52 \\\hline
\end{tabular}
\label{tab:table7}
\end{table}

\begin{table*}[!htb]
\caption{YOLOv7 mAP@0.5 Scores (For All Classes)}
\small 
\setlength\tabcolsep{3.5pt}
    \begin{tabular}{l c c c c c c c c c}
    \hline
    Model variant & Boneanomality & Bonelesion & Foreignbody & Fracture & Metal & Periostealreaction & Pronatorsign & Softtissue & Text\\\hline
YOLOv7-Tiny & 0.13 & 0.00 & 0.00 & 0.93 & 0.88 & 0.69 & 0.69 & 0.14 & 0.99\\
YOLOv7 & 0.20 & 0.33 & 0.33 & 0.94 & 0.95 & 0.76 & 0.71 & 0.25 & 0.99\\
YOLOv7x & 0.17 & 0.10 & 0.00 & 0.94 & 0.90 & 0.72 & 0.70 & 0.24 & 0.99\\
YOLOv7-W6 & 0.00 & 0.00 & 0.00 & 0.90 & 0.88 & 0.57 & 0.46 & 0.14 & 0.98\\
YOLOv7-E6 & 0.00 & 0.00 & 0.00 & 0.83 & 0.81 & 0.42 & 0.40 & 0.11 & 0.98\\
YOLOv7-D6 & 0.00 & 0.00 & 0.00 & 0.89 & 0.87 & 0.53 & 0.34 & 0.14 & 0.99\\
YOLOv7-E6E & 0.01 & 0.00 & 0.00 & 0.91 & 0.88 & 0.60 & 0.43 & 0.12 & 0.99\\\hline
\label{tab:table9}
\end{tabular}
\end{table*}

Tables \ref{tab:table11} and \ref{tab:table12} show the performance of YOLOv8 model variants across all classes and on the fracture class, respectively. The YOLOv8 variant "YOLOv8x" achieved the highest mAP of 0.95 for fracture detection at an IoU threshold of 0.5, with a precision of 0.91 and a recall of 0.89. Additionally, it demonstrated superior overall performance across all classes with a mAP of 0.77 at an IoU threshold of 0.5. Table \ref{tab:table13} also shows that all YOLOv8 variants demonstrated good performance in detecting all classes, including minority classes, except the "foreign body" class not being detected by the small and the medium variants. The results suggest that using compound-scaled variants of the YOLOv8 architecture generally improves performance, except for a decrease in mAP scores across all classes when moving from the variant "s" to a medium variant "m", with a decrease of 0.09 in Table \ref{tab:table12}.

\begin{table}[htbp]
\caption{YOLOv8 Results (Across All Classes)}
\small 
\setlength\tabcolsep{4pt} 
\begin{tabular}{@{}lccccc@{}}
\hline
    Model variant & Precision & Recall & mAP@0.5 & mAP@0.5:0.95\\
    \hline
YOLOv8n & 0.73 & 0.58 & 0.59 & 0.36 \\
YOLOv8s & 0.72 & 0.63 & 0.65 & 0.39 \\
YOLOv8m & 0.60 & 0.60 & 0.56 & 0.36 \\
YOLOv8l & 0.74 & 0.60 & 0.62 & 0.41 \\
YOLOv8x & 0.79 & 0.64 & 0.77 & 0.53 \\\hline
\label{tab:table11}
\end{tabular}
\end{table}

\begin{table}[htbp]
\caption{YOLOv8 Results (Fracture Class)}
\small 
\setlength\tabcolsep{4pt} 
\begin{tabular}{@{}lccccc@{}}
\hline
    Model variant & Precision & Recall & mAP@0.5 & mAP@0.5:0.95\\
    \hline
YOLOv8n & 0.87 & 0.88 & 0.93 & 0.55 \\
YOLO8s & 0.87 & 0.91 & 0.94 & 0.56 \\
YOLOv8m & 0.84 & 0.92 & 0.95 & 0.57 \\
YOLOv8l & 0.92 & 0.90 & 0.95 & 0.57 \\
YOLO8x & 0.91 & 0.89 & 0.95 & 0.57 \\\hline
\label{tab:table12}
\end{tabular}
\end{table}

\begin{table*}[!htb]
\caption{YOLOv8 mAP@0.5 Scores (For All Classes)}
\small 
\setlength\tabcolsep{3.5pt}
    \begin{tabular}{l c c c c c c c c c}
    \hline
    Model variant & Boneanomality & Bonelesion & Foreignbody & Fracture & Metal & Periostealreaction & Pronatorsign & Softtissue & Text\\
    \hline
YOLO8n & 0.20 & 0.50 & 0.11 & 0.93 & 0.91 & 0.71 & 0.70 & 0.26 & 0.99 \\
YOLOv8s & 0.27 & 1.00 & 0.00 & 0.94 & 0.93 & 0.71 & 0.76 & 0.21 & 0.99 \\
YOLOv8m & 0.19 & 0.27 & 0.00 & 0.95 & 0.96 & 0.73 & 0.80 & 0.18 & 0.99 \\
YOLOv8l & 0.22 & 0.55 & 0.10 & 0.95 & 0.97 & 0.72 & 0.79 & 0.26 & 0.99 \\
YOLOv8x & 0.26 & 1.00 & 1.00 & 0.95 & 0.95 & 0.72 & 0.79 & 0.32 & 0.99 \\
\hline
\label{tab:table13}
\end{tabular}
\end{table*}

\begin{figure}
\includegraphics[width=0.98\linewidth, height=7cm]{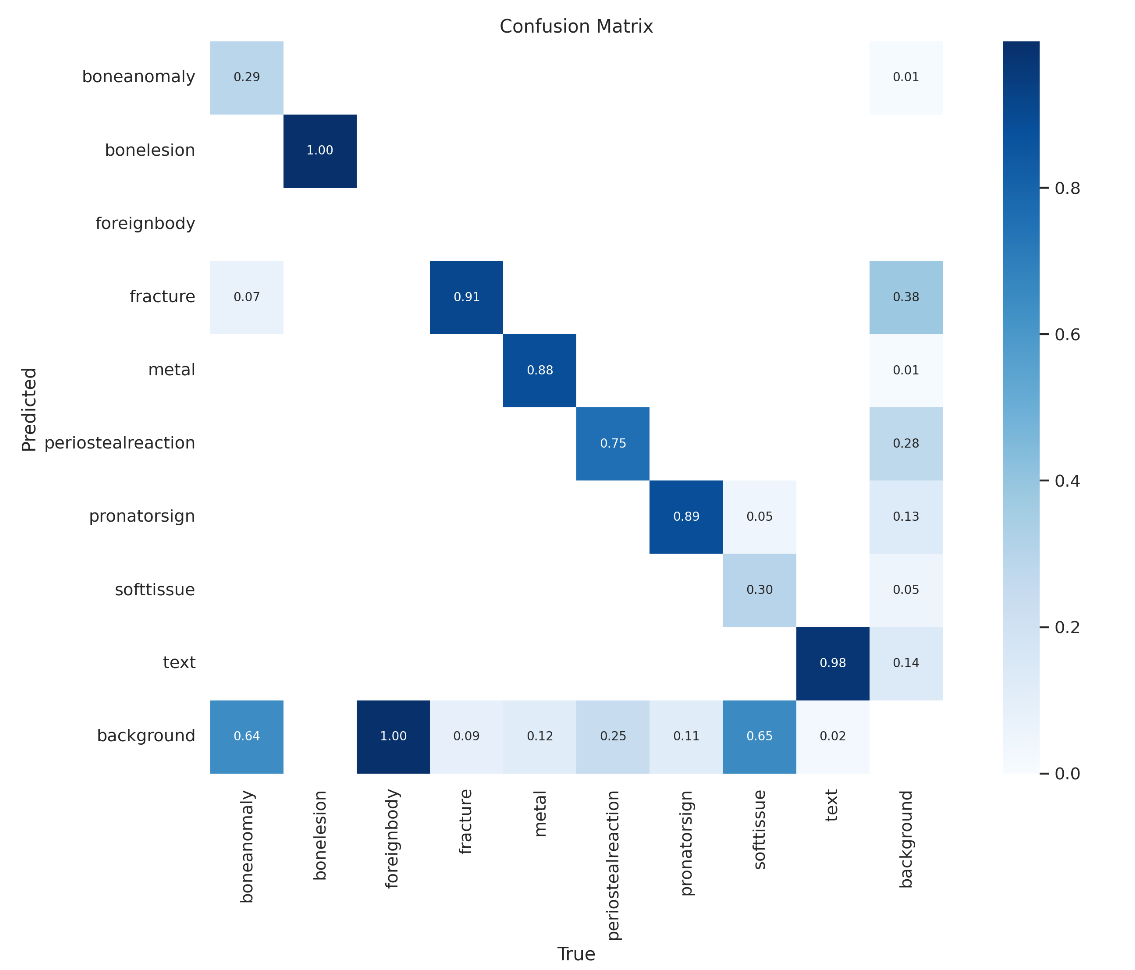}
\caption{{Confusion Matrix (YOLOv8x)}.}
\label{fig1:confmat_v8}
\end{figure}

The results of the experimental evaluation using the two-stage detector Faster R-CNN are presented in Table \ref{tab:table10}. The table shows the mean Average Precision (mAP) scores obtained for each class individually as well as the overall mAP across all classes. The results indicate that all variants of the YOLO model outperform Faster R-CNN by a significant margin. This is supported by the fact that the mean mAP score of every YOLO variant was found to be higher than that of Faster R-CNN, both for fracture detection and overall performance across all classes. These findings suggest that the single-stage detection algorithm, YOLO, is a more effective model for this task. Moreover, Faster R-CNN does not seem to exhibit the ability to detect the classes in minority such as "bone anomaly", "bone lesion", and "foreign body".

\begin{table}[!htb]
    \centering
\caption{Faster R-CNN mAP@0.5 Scores (Across All Classes}
    \begin{tabular}{l r }
    \hline
    Abnormality  & mAP@0.5 \\
    \hline
Boneanomaly & 0.00\\
Bonelesion & 0.00 \\
Foreignbody & 0.00\\
Fracture & 0.75\\
Metal & 0.78 \\
Periostealreaction & 0.54\\
Pronatorsign & 0.10\\
Softtissue & 0.03\\
Text & 0.96\\
All & 0.35\\
\hline
\end{tabular}
\label{tab:table10}
\end{table}

\begin{figure*}
\centering
\includegraphics[width=0.7\textwidth]{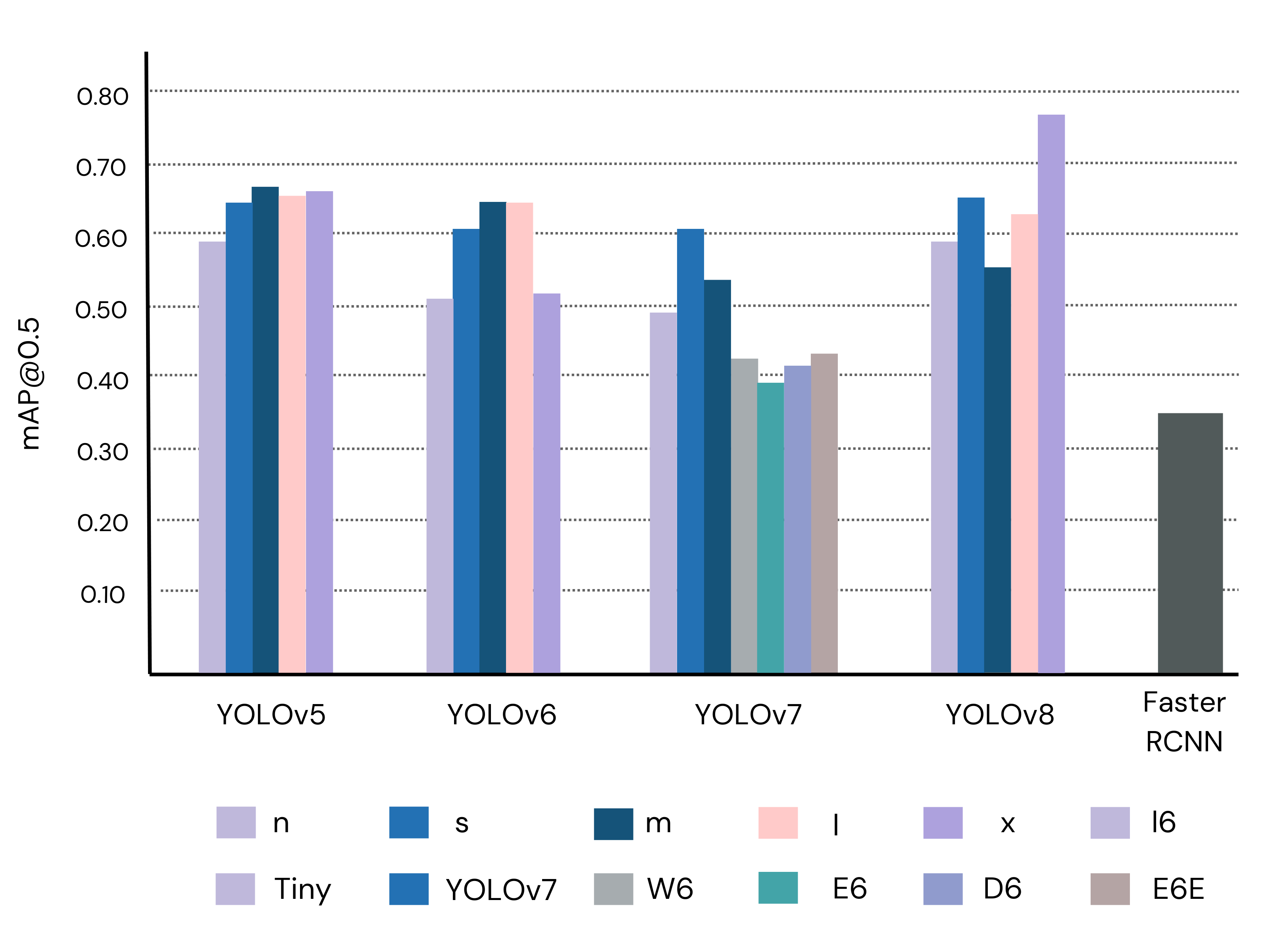}
\caption{{mAP Scores (Across All Classes)}.}
\label{fig1:mapall}
\end{figure*}

\begin{figure*}
\centering
\includegraphics[width=0.7\textwidth]{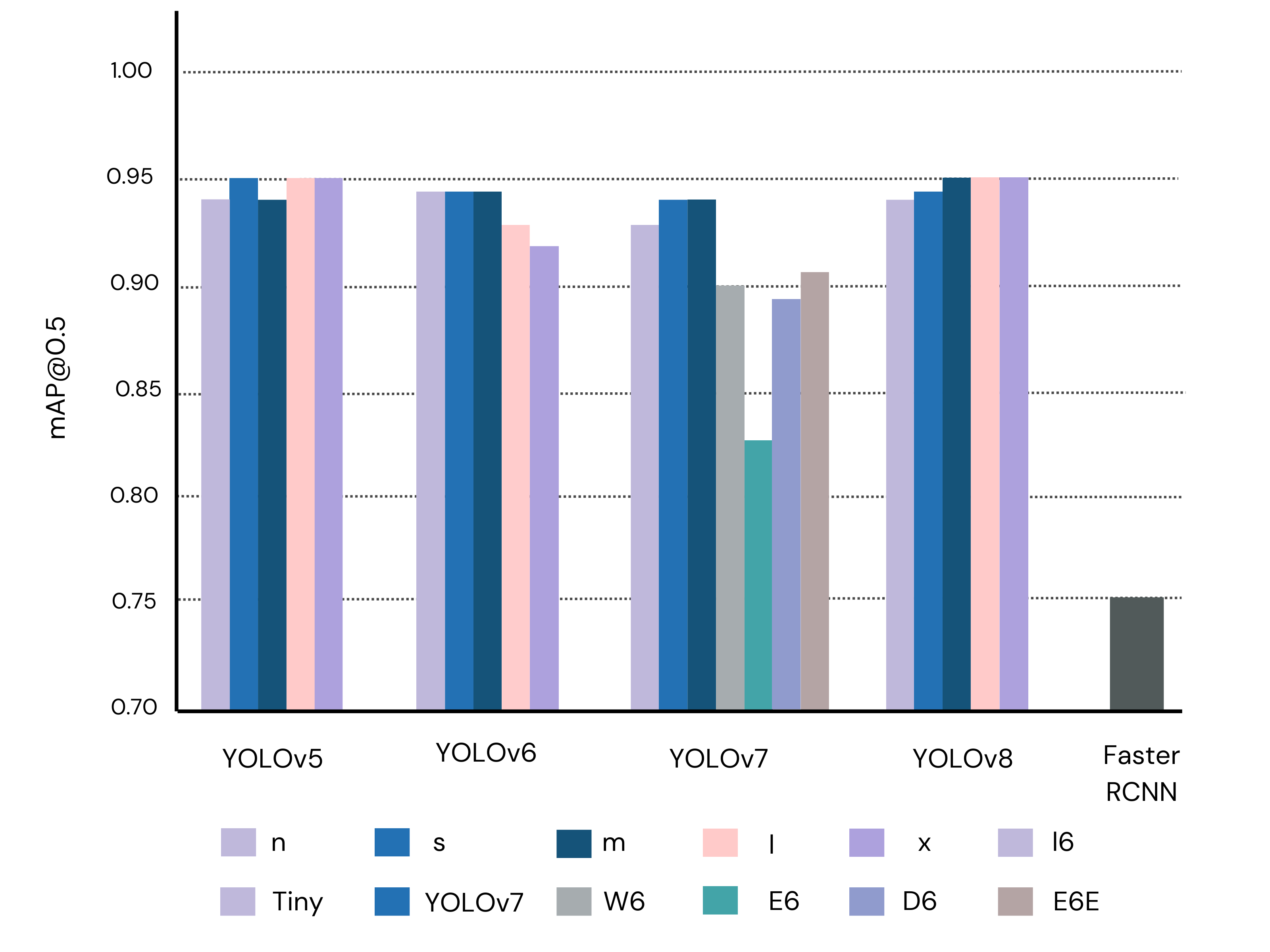}
\caption{{mAP Scores (Fracture Class)}.}
\label{fig1:mapfrac}
\end{figure*}

\begin{figure*}
\includegraphics[width=1\textwidth, height=11cm]{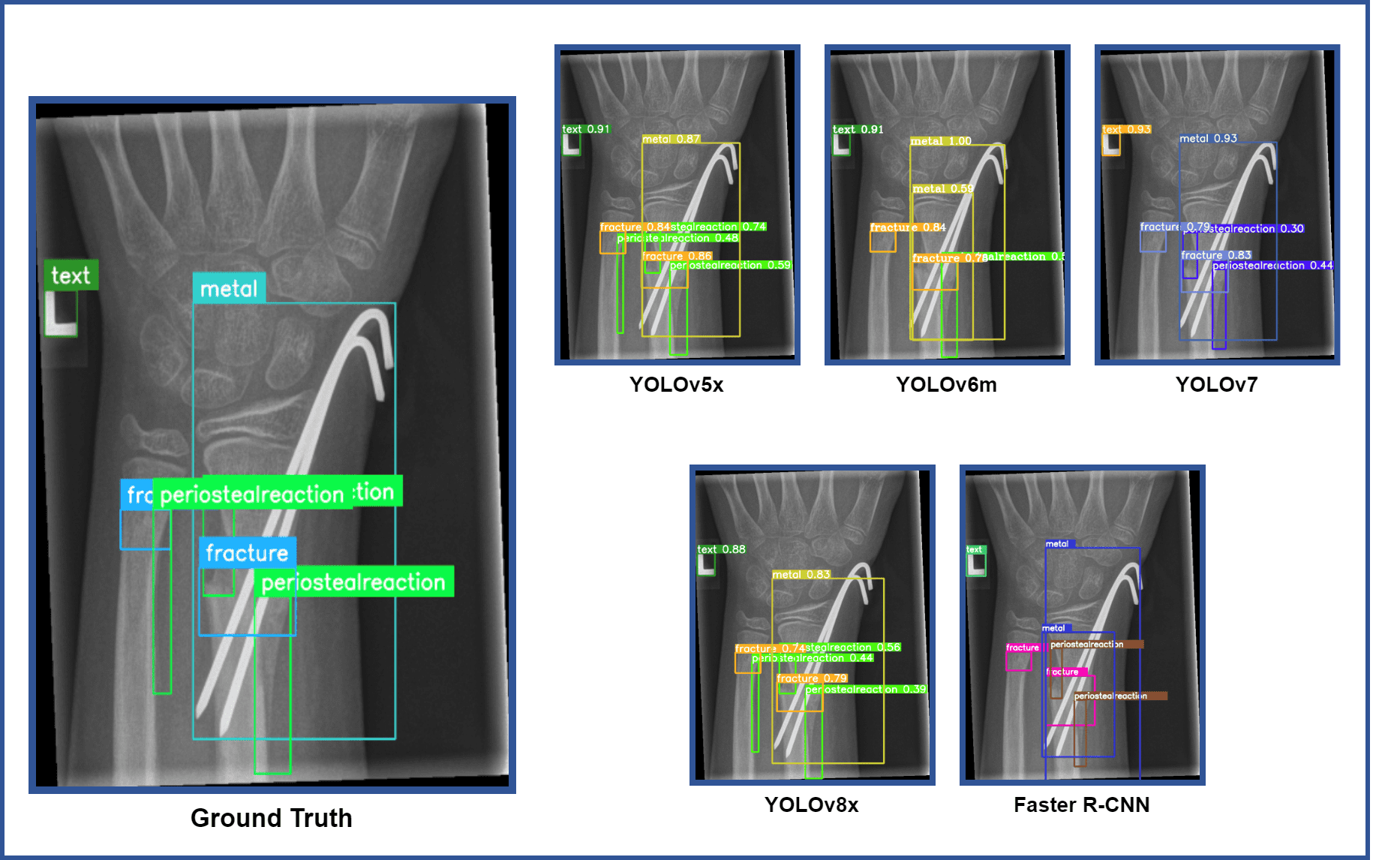}
\caption{Bounding box estimated by YOLO variants and Faster R-CNN.}

\label{fig1:inference}
\end{figure*}

Figures \ref{fig1:mapall} and \ref{fig1:mapfrac} provide an overview of the mAP scores obtained for fracture class as well as across all classes by all YOLO variants and Faster R-CNN.
In applications where false positives are costly, a model with high precision may be preferable, while in situations where missing detections are costly, a model with high recall may be more desirable. The mean Average Precision (mAP) serves as a comprehensive measure of the model's performance. Therefore, we selected the best-performing variant within each YOLO model based on the highest mAP achieved for the fracture class and overall performance across all classes. We have also compared their mAP scores to each other as well as with that of the Faster R-CNN model, as illustrated in Table \ref{tab:table17}. We also evaluated the performance of all variants, including Faster R-CNN, on a challenging image containing multiple objects of interest, including 2 fractures, 3 periosteal reactions, 1 metal, and 1 text. The bounding box estimates for these objects from each variant and Faster R-CNN are illustrated in Fig. \ref{fig1:inference}. 

It is clear from Table \ref{tab:table17} that the variant "YOLOv8x" of YOLOv8 is the best-performing variant out of all the variants employed in this study. The results presented in this study using the variant "YOLOv8x" represent a significant improvement upon the ones originally presented in \cite{Nagy2022} for the fracture class. In that paper, the model variant "YOLOv5m" trained on COCO weights achieved a mean average precision (mAP) score of 0.93 for fracture detection and an overall mAP score of 0.62 at an IoU threshold of 0.5. In contrast, the results obtained in this study demonstrate a higher mAP score of 0.95 for fracture detection and an overall mAP of 0.77 at an IoU threshold of 0.5. Fig. \ref{f1fig1}, \ref{precfig1}, \ref{recfig1}, and \ref{prfig1} present the F1 versus Confidence, Recall versus Confidence, Precision versus Confidence, and Precision versus Recall curves, respectively, for the variant "YOLOv8x" across all classes. These curves provide a visual representation of the model's performance on different confidence intervals and allow for a more thorough evaluation of its capabilities. The F1 versus Confidence curve shows the relationship between the model's F1 score, which is a measure of the balance between precision and recall, and the confidence of its predictions. The Recall versus Confidence curve illustrates the model's ability to correctly identify objects, while the Precision versus Confidence curve demonstrates the proportion of correct predictions made by the model. The Precision versus Recall curve shows the trade-off between the model's precision and recall, with higher precision typically corresponding to lower recall and vice versa. Additionally, a confusion matrix \ref{fig1:confmat_v8} is shown for the variant "YOLOv8x".

\begin{table}[!h]
    \centering
\caption{mAP@0.5 Scores For Best Performing Model Variants}
    \begin{tabular}{l c c}
    \hline
    Model  & Fracture & All \\
    \hline
YOLOv5x & 0.95 & 0.69\\
YOLOv6m & 0.94 & 0.64 \\
YOLOv7 & 0.94 & 0.61\\
YOLOv8x & \textbf{0.95} & \textbf{0.77}\\
Faster R-CNN & 0.75 & 0.35\\
\hline
\end{tabular}
\label{tab:table17}
\end{table}

Our study found that the relationship between the complexity of a YOLO model and its performance is not always linear. Our results on the GRAZPEDWRI-DX dataset revealed that the performance of YOLO models did not consistently improve with increasing complexity, except for YOLOv5 and YOLOv8. 

\section{Conclusion \& Future Work}
In this study, we aimed to evaluate the performance of state-of-the-art single-stage detection models, specifically YOLOv5, YOLOv6, YOLOv7, and YOLOv8, in detecting wrist abnormalities and compare their performances against each other and the widely used two-stage detection model Faster R-CNN. Additionally, the analysis of the performance of all variants within each YOLO model was also provided. The evaluation was conducted using the recently released GRAZPEDWRI-DX \cite{Nagy2022} dataset, with a total of 23 detection procedures being carried out. The findings of our study demonstrated that YOLO models outperform the commonly used two-stage detection model, Faster R-CNN, in both fracture detection and across all classes present in the GRAZPEDWRI-DX dataset. 

Furthermore, an analysis of YOLO models revealed that the YOLOv8 variant "YOLOv8x" achieved the highest mAP across all classes of wrist abnormalities in the GRAZPEDWRI-DX dataset, including the fracture class, at an IoU threshold of 0.5. We also discovered that the relationship between the complexity of a YOLO model, as measured by the use of compound-scaled variants within each YOLO model, and its performance is not always linear. Specifically, our analysis of the GRAZPEDWRI-DX dataset revealed that the performance of YOLO variants did not consistently improve with increasing complexity, except for YOLOv5 and YOLOv8. Some variants were successful in detecting minority classes while others were not. These results contribute to understanding the relationship between the complexity of YOLO models and their performance, which is important for guiding the development of future models. Our study highlights the potential of single-stage detection algorithms, specifically YOLOv5, YOLOv6, YOLOv7, and YOLOv8, for detecting wrist abnormalities in clinical settings. These algorithms are faster than their two-stage counterparts, making them more practical for emergencies commonly found in hospitals and clinics. Additionally, the study's results indicate that single-stage detectors are highly accurate in detecting wrist abnormalities, making them a promising choice for clinical use. 

While this research was conducted, YOLOv8 was the most recent version. The results of this study can serve as a benchmark for evaluating the performance of future models for wrist abnormality detection, as further improvements to either YOLOv8 or future versions of YOLO may surpass the results obtained in this study. It is worth noting that this study didn't explore the entire hyperparameter space and finding the best hyperparameters for each YOLO model may improve wrist abnormality detection performance on the dataset. Computational limitations restricted the input resolution to 640 pixels, but higher resolutions could further improve performance. The study showed that the models had difficulty detecting "bone anomaly", "bone lesion", and "foreign body" due to low instances of these classes, so increasing their instances through augmentation or image generation could enhance performance. Additionally, the performance of classification models could also be assessed by exploring the dataset for pure classification tasks without object localization.

{\color{black}
\section{Acknowledgement}
}
{\color{black}
This work was supported in part by the Department of Computer Science (IDI), Faculty of Information Technology and Electrical Engineering, Norwegian University of Science and Technology (NTNU), Gj\o vik, Norway; and in part by the Curricula Development and Capacity Building in Applied Computer Science for Pakistani Higher Education Institutions (CONNECT) Project NORPART-2021/10502, funded by DIKU.
}

\bibliographystyle{cas-model2-names}

\bibliography{cas-refs}

\end{document}